%% file: arxiv.tex
\documentclass{article}

\usepackage{enumitem}
\usepackage{url}
\usepackage{hyperref}

\usepackage{spconf,amsmath,graphicx}
\usepackage{amssymb}

\date{}

\title{RADMI: LATENT INFORMATION AGGREGATION AS A PROXY FOR MODEL UNCERTAINTY}

\name{William Stevens, Mohit Prabhushankar, and Ghassan AlRegib\thanks{This work is supported by the ML4Seismic Consortium at Georgia Tech.}}
\address{OLIVES at the Center for Signal and Information Processing,\\
School of Electrical and Computer Engineering, Georgia Institute of Technology, Atlanta, GA, USA\\
\{wstevens35, mohit.p, alregib\}@gatech.edu}

\begin{document}

% ============================================================================
% COVER PAGE
% ============================================================================
\onecolumn
\thispagestyle{empty}

\begin{description}[labelindent=0cm,leftmargin=3cm,style=multiline,itemsep=3ex]

\item[\textbf{Citation}]{W. Stevens, M. Prabhushankar, and G. AlRegib, ``RADMI: Latent Information Aggregation as a Proxy for Model Uncertainty,'' in \textit{IEEE International Conference on Image Processing (ICIP)}, Tampere, Finland, 2026.}

\item[\textbf{Review}]{Date of Acceptance: April 30th 2026}

\item[\textbf{Codes}]{\url{https://github.com/olivesgatech/RADMI}}

\item[\textbf{Bib}]{@inproceedings\{stevens2026radmi,\\
  title=\{RADMI: Latent Information Aggregation as a Proxy for Model Uncertainty\},\\
  author=\{Stevens, William and Prabhushankar, Mohit and AlRegib, Ghassan\},\\
  booktitle=\{IEEE International Conference on Image Processing (ICIP)\},\\
  year=\{2026\},\\
  location=\{Tampere, Finland\}\}}

\item[\textbf{Copyright}]{\textcopyright~2026 IEEE. Personal use of this material is permitted. Permission from IEEE must be obtained for all other uses, in any current or future media, including reprinting/republishing this material for advertising or promotional purposes, creating new collective works, for resale or redistribution to servers or lists, or reuse of any copyrighted component of this work in other works.}

\item[\textbf{Contact}]{\href{mailto:alregib@gatech.edu}{alregib@gatech.edu}\\
\url{https://alregib.ece.gatech.edu/}}

\end{description}

\newpage
\setcounter{page}{1}

% ============================================================================
% PAPER CONTENT
% ============================================================================
\twocolumn
\ninept

\maketitle

\begin{abstract}

\input{sections/01-abstract}
\end{abstract}

\section{Introduction}
\label{sec:intro}

\input{sections/02-intro}

\vspace{-2mm}
\section{Related Works}
\label{sec:relatedworks}

\input{sections/03-related-works}

\vspace{-2mm}
\section{Methodology}
\label{sec:methodology}
\input{sections/04-methodology}

\vspace{-2mm}
\section{Experiments and Results}
\label{sec:experiments}
\input{sections/05-experiments}

\vspace{-2mm}
\section{Conclusion}
\label{sec:conclusion}
\input{sections/06-conclusion}

\vfill
\pagebreak

% ============================================================================
% REFERENCES
% NOTE: For arXiv, replace \bibliography{refs} with contents of your .bbl file
% ============================================================================

\end{document}

%% file: sections/01-abstract.tex
Epistemic uncertainty estimation is essential for identifying regions where deep learning system outputs may be unreliable. However, existing approaches require computationally expensive ensemble methods or multiple stochastic forward passes, limiting their scalability to dense prediction tasks like segmentation. We propose Resolution-Aggregated Decoder Mutual Information (\texttt{RADMI}), a single-pass method that estimates prediction uncertainty by measuring mutual information (MI) between consecutive decoder layers in segmentation networks. We observe that elevated inter-layer MI correlates with prediction uncertainty, as the network must integrate conflicting contextual information at ambiguous regions such as class boundaries. Evaluating on a seismic facies segmentation benchmark, \texttt{RADMI} achieves the highest correlation with deep ensemble uncertainty among all single-pass methods, outperforming the next-best baselines by 5.5\% in Pearson and 10.7\% in Spearman correlation coefficients. Compared to baselines that either lack spatial precision or demand significant computational overhead, \texttt{RADMI} yields sharp, boundary-localized uncertainty maps without architectural modifications. Our results suggest that linear aggregation of normalized information flow provides a principled and efficient proxy for prediction uncertainty in encoder-decoder architectures.

\begin{keywords}
Deep Learning, Uncertainty Estimation, Mutual Information, Seismic Interpretation, Segmentation
\end{keywords}

%% file: sections/02-intro.tex
Deep learning has found success in automating seismic interpretation tasks, including facies classification \cite{alaudahfacies}, fault delineation \cite{quesada2025large}, and salt body segmentation \cite{alregib2018subsurface}. By learning hierarchical feature representations directly from migrated seismic volumes, neural networks can produce interpretations that approach expert-level quality on well-characterized geological formations. However, deep networks are prone to overconfident predictions, even in regions where their outputs are unreliable \cite{guo2017calibration}. %For seismic data, surveys produce large 3D scans of the Earth's crust. 
In seismic interpretation workflows, a small section of the 3D seismic survey is annotated by a human expert, while the rest of the volume's interpretation is handled by a deep learning network. This automated labeling carries inherent noise and uncertainty, that may lead to biased interpretation~\cite{chowdhury2025unified}.

Segmentation uncertainty quantification addresses this challenge by providing pixel-wise confidence estimates alongside segmentation outputs. In the context of deep learning, uncertainty can be broadly categorized into aleatoric uncertainty, arising from inherent noise in the data (such as labeling ambiguity), and epistemic uncertainty, arising from limited training data or model capacity~\cite{kendall2017uncertainties}. In principle, if a model can identify which regions it is uncertain about, interpreters can focus manual review on these areas while trusting the automated predictions elsewhere~\cite{benkert2024effective}. Existing approaches to uncertainty estimation in seismic interpretation predominantly rely on Bayesian techniques \cite{zhaointerpretation} \cite{choiuncertainty}. Deep ensembles aggregate predictions from multiple independently trained models \cite{lakshminarayanan2017simple}. Monte Carlo Dropout (MCD) performs multiple stochastic forward passes with dropout enabled at test time \cite{galpmlr}. Prediction switch tracking counts classification changes during training as a measure of learning difficulty \cite{benkert2022reliableuncertaintyestimation}. While effective, these methods require either multiple trained models, multiple forward passes, or access to training dynamics unavailable for new data. For large-scale seismic volumes spanning hundreds of square kilometers worth of data, this computational overhead poses a significant barrier to deployment. %Encoder-decoder architectures such as U-Net \cite{ronneberger2015unetconvolutionalnetworksbiomedical} have become standard for seismic segmentation due to their ability to capture both global context and fine-grained spatial detail. %Our base architecture, FaciesSegNet~\cite{}, is a U-Net variant optimized for seismic facies classification that we use throughout this work. However, 
%Our proposed method is architecture-agnostic and can be applied to any encoder-decoder network.

\begin{figure}[h]
\begin{minipage}[b]{1.0\linewidth}
  \centering
  \centerline{\includegraphics[width=8.5cm]{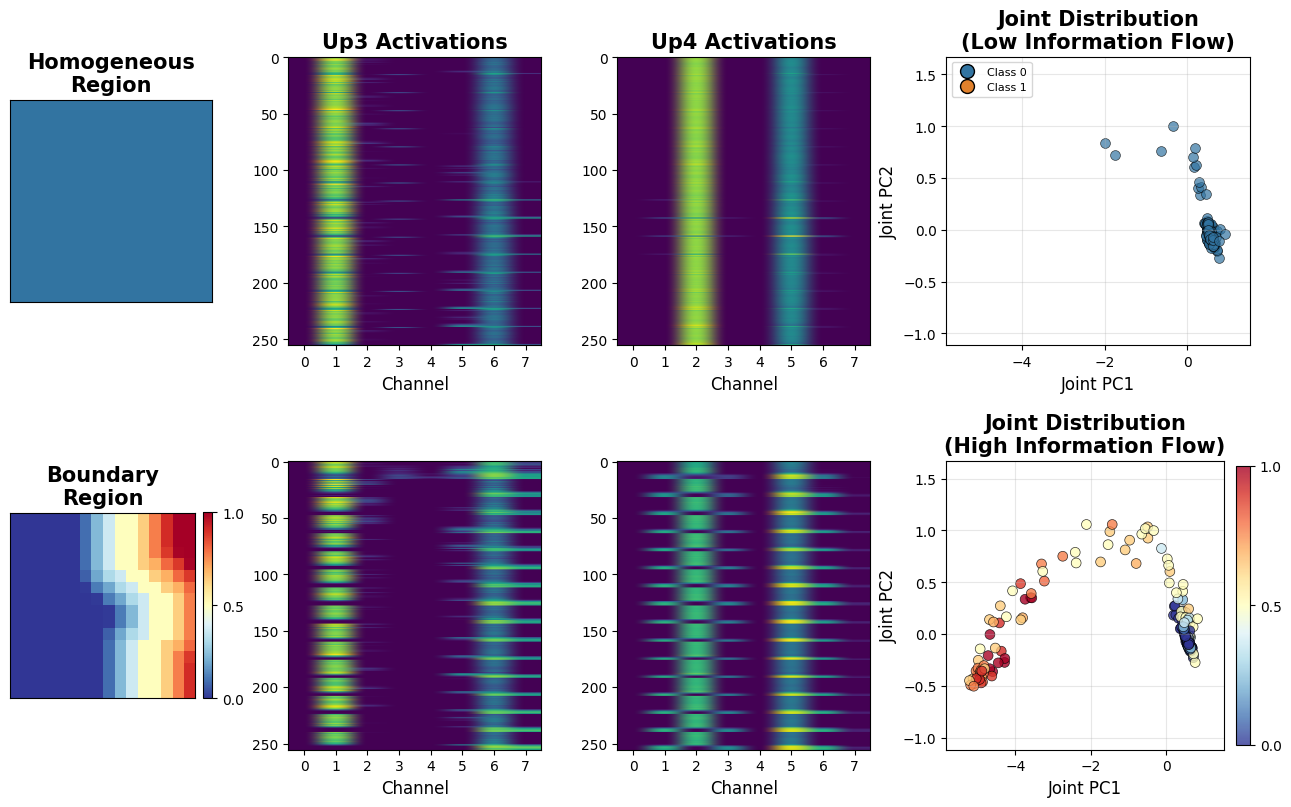}}
%  \vspace{2.0cm}
\end{minipage}
\vspace{-5mm}
\caption{Homogeneous regions (top) yield tightly clustered decoder activations and low MI. Class boundaries (bottom) produce spread distributions and elevated MI, signaling uncertainty.}
\label{fig:intuition}
\end{figure}

In this paper, we propose Resolution-Aggregated Decoder Mutual Information (\texttt{RADMI}), a novel single-pass uncertainty estimation method based on information flow within any given encoder-decoder architecture. The U-Net architecture~\cite{ronneberger2015unetconvolutionalnetworksbiomedical} and its variants like the FaciesSegNet~\cite{mustafa2021man} are commonly used for facies segmentation. Our key observation is that the mutual information (MI) between consecutive decoder layers reflects the network's internal processing demands during a forward segmentation pass. %MI quantifies the statistical dependence between two random variables, in this case: the feature representations at adjacent decoder stages. 
Figure~\ref{fig:intuition} illustrates this phenomenon. In homogeneous regions activations cluster together, resulting in low inter-layer MI. At class boundaries, the decoder must reconcile conflicting information from neighboring spatial contexts, producing a spread distribution with stronger statistical dependence and elevated MI. By measuring this dependence, we obtain a principled uncertainty estimate from a single forward pass. We validate \texttt{RADMI} on the Netherlands F3 block seismic facies benchmark \cite{alaudahfacies}, comparing against established baselines including softmax entropy~\cite{hendrycks2016baseline}, maximum softmax probability~\cite{hendrycks2016baseline}, MC-Dropout~\cite{galpmlr}, deep ensembles~\cite{lakshminarayanan2017simple}, and prediction switches~\cite{benkert2022reliableuncertaintyestimation}.

%% file: sections/03-related-works.tex
\subsection{Uncertainty Quantification in Semantic Segmentation}
Uncertainty quantification for semantic segmentation presents unique challenges beyond image classification, as uncertainty must be estimated at the pixel level rather than for the semantics of the entire image. The most common single-pass approaches derive uncertainty from the softmax output distribution. Predictive entropy measures the spread of the categorical distribution, while maximum softmax probability (MSP) uses confidence in the top prediction as an inverse uncertainty measure~\cite{hendrycks2016baseline}. These methods are computationally efficient but are known to produce overconfident estimates, particularly for out-of-distribution inputs~\cite{prabhushankar2024counterfactual}.

Two fundamental problems exist in uncertainty estimation approaches. The first is output calibration, where neural networks produce overconfident predictions that do not reflect true uncertainty~\cite{guo2017calibration}. Softmax-based methods inherit this limitation, as they operate directly on potentially miscalibrated output distributions. The second is computational cost. Methods that address calibration, such as deep ensembles~\cite{lakshminarayanan2017simple} and Monte Carlo Dropout (MCD)~\cite{galpmlr}, require multiple trained models or multiple forward passes respectively. For applications involving large-scale data such as 3D seismic volumes, this overhead poses a significant barrier to deployment. Prediction switches, which count the frequency of each pixel's classification change across training epochs, are proposed as a measure of learning difficulty~\cite{benkert2022reliableuncertaintyestimation}. However, this approach requires saving model checkpoints at every epoch and running inference through all of them for new data, resulting in hundreds of forward passes per sample. \texttt{RADMI} addresses both problems simultaneously, providing well-localized uncertainty estimates in a single forward pass without requiring architectural modifications or ensemble training.

\subsection{Evaluating Uncertainty}

Evaluating uncertainty estimation methods is inherently challenging since there is no ground-truth uncertainty. In natural image domains, existing techniques indirectly measure uncertainty quality through downstream applications such as out-of-distribution detection~\cite{hendrycks2016baseline}, active learning~\cite{benkert2024effective}, and sensitivity analysis~\cite{prabhushankar2024voice}. However, downstream evaluation for seismic interpretation requires human-in-the-loop frameworks involving expert geologists, which is not scalable.

In this work, we treat deep ensemble uncertainty as the reference standard for evaluation. Deep ensembles provide a principled approximation of epistemic uncertainty by sampling from the space of plausible models through independent training with different random initializations~\cite{lakshminarayanan2017simple}. Unlike MCD, which approximates a restricted posterior through dropout perturbations, ensembles make minimal assumptions about the posterior distribution. %Empirically, ensembles consistently outperform other Bayesian approximations across diverse benchmarks~\cite{lakshminarayanan2017simple}. 
Hence, any single-pass method that approximates ensemble uncertainty provides practical value for real-world deployment. % without sacrificing the quality of uncertainty estimates.

\subsection{Information Theory in Deep Networks}
The Information Bottleneck (IB) principle \cite{tishby2000informationbottleneckmethod}
provides a theoretical framework for characterizing optimal representations $Z$ within a neural network. For a given representation $Z$ and target labels $Y$, IB states that training maximizes the mutual information $I(Z; Y)$ between the representation and target. Hence, knowing $Y$ reduces the uncertainty in $Z$ and vice versa. Note that maximizing $I(Z; Y)$ is subject to the constraint of minimizing the mutual information between the input $X$ and the representation $Z$, $I(X; Z)$. Qualitatively, the network must compress confounding features in $X$ to extract the minimal sufficient statistics in $Z$ that allow predicting $Y$. %Subsequent work \cite{tishby2015deeplearninginformationbottleneck} \cite{shwartzziv2017openingblackboxdeep}, used mutual information to analyze the evolution of the representation space during training, revealing distinct phases of fitting and compression.
%MI and uncertainty share a fundamental connection. 
%However, regions where the network is uncertain correspond to inputs that are difficult to compress into a sufficient representation. At class boundaries, conflicting information from neighboring regions increases the complexity of the representation, which manifests as elevated MI between processing stages. %However, the standard IB formulation measures $I(T; Y)$, the mutual information between layer representations $Z$ and the target labels $Y$, requiring labeled data that is unavailable at inference time. 
Since $Y$ is unavailable at inference, the reduction in uncertainty, $I(Z; Y)$ cannot be directly calculated. \texttt{RADMI} bridges this gap by measuring mutual information between consecutive layer representations, eliminating the dependence on labels while capturing meaningful structure about the network's internal processing and information flow.

Prior information-theoretic analyses of deep networks have focused on input-to-representation compression $I(X; Z)$ or representation-to-output prediction $I(Z; Y)$~\cite{shwartzziv2017openingblackboxdeep, tishby2015deeplearninginformationbottleneck}, leaving inter-layer information flow $I(Z_l; Z_{l+1})$ underexplored to the best of our knowledge. In segmentation decoders, successive layers must propagate both semantic and spatial information; at class boundaries, these conflict, increasing inter-layer dependence. \texttt{RADMI} exploits this by measuring MI between consecutive decoder layers, capturing uncertainty without labels or multiple forward passes.

%% file: sections/04-methodology.tex
\begin{figure*}[htb]
\vspace{-10mm}
\begin{minipage}[b]{1.0\linewidth}
  \centering
  \centerline{\includegraphics[width=17cm]{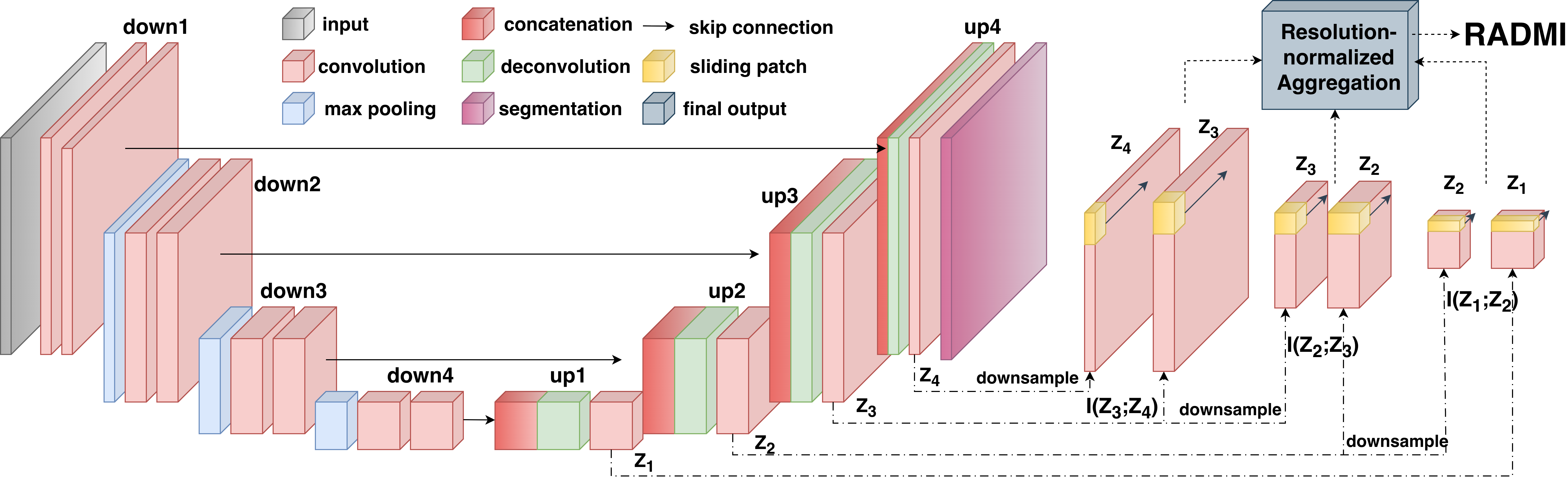}}
%  \vspace{2.0cm}
\end{minipage}
\vspace{-5mm}
\caption{\texttt{RADMI} Workflow.}
\label{fig:method}
\end{figure*}

Figure~\ref{fig:method} illustrates the \texttt{RADMI} workflow. We propose a single-pass uncertainty estimation method based on measuring mutual information between consecutive decoder layers in segmentation networks. The core methodology measures the mutual information
\begin{equation}
    I(\mathbf{Z}_l; \mathbf{Z}_{l+1})
\end{equation}
which quantifies the statistical dependence between adjacent decoder stages $l$ and $l+1$. Unlike methods requiring predictive distributions or label access, this formulation operates purely on internal activations during a single forward pass.

The intuition follows directly from the relationship between MI and uncertainty established in Section \ref{sec:relatedworks}. At class boundaries, the decoder must integrate information from neighboring regions belonging to different semantic categories. This integration increases the statistical dependence between consecutive layer representations, as each layer's activations become more predictive of the next layer's response. By measuring this inter-layer dependence, we obtain a proxy for prediction uncertainty without requiring labels or multiple forward passes.

\subsection{Spatially-Localized MI Estimation}

To generate pixel-level uncertainty maps, we compute MI locally using a sliding window approach. Given feature maps from consecutive layers with potentially different spatial resolutions, we first align them via bilinear interpolation to the coarser resolution. For each spatial location $(i,j)$, we extract corresponding $p \times p$ patches from both aligned feature maps.

Within each patch, we treat the $p^2$ spatial positions as samples drawn from a joint distribution over channel activations. Let $\mathbf{a}_k \in \mathbb{R}^{C}$ and $\mathbf{b}_k \in \mathbb{R}^{C}$ denote the $C$-dimensional channel vectors at spatial position $k$ within aligned patches from layers $l$ and $l+1$ respectively. We model these as jointly Gaussian:
\begin{equation}
    \begin{bmatrix} \mathbf{a} \\ \mathbf{b} \end{bmatrix} \sim \mathcal{N}\left(\boldsymbol{\mu}, \boldsymbol{\Sigma}\right), \quad \boldsymbol{\Sigma} = \begin{bmatrix} \boldsymbol{\Sigma}_{\mathbf{a}} & \boldsymbol{\Sigma}_{\mathbf{ab}} \\ \boldsymbol{\Sigma}_{\mathbf{ab}}^\top & \boldsymbol{\Sigma}_{\mathbf{b}} \end{bmatrix}
\end{equation}

Under the Gaussian assumption, MI has the following closed-form expression~\cite{cover2006elements}:
\begin{equation}
    I(\mathbf{A}; \mathbf{B}) = \frac{1}{2}\log\frac{|\boldsymbol{\Sigma}_{\mathbf{a}}||\boldsymbol{\Sigma}_{\mathbf{b}}|}{|\boldsymbol{\Sigma}|}
\end{equation}
Covariance matrices are estimated empirically from the $p^2$ samples within each patch. This procedure yields an MI value for each patch location. Using stride $s=1$, we obtain a dense MI map at the aligned resolution.

\subsection{Resolution-Weighted Aggregation}

For a decoder with $L$ upsampling stages, we compute mutual information maps between all consecutive layer pairs:
\[\{(\mathbf{Z}_1, \mathbf{Z}_2), \ldots, (\mathbf{Z}_{L-1}, \mathbf{Z}_L)\}\]
Each pair produces an MI map computed at the coarser layer's resolution before upsampling to the output dimensions.

For a decoder with $L$ upsampling stages, we compute MI maps between all consecutive layer pairs $\{(\mathbf{Z}_1, \mathbf{Z}_2), \ldots, (\mathbf{Z}_{L-1}, \mathbf{Z}_L)\}$. Each MI map $\mathbf{M}_l$ is computed at the coarser layer's resolution $H_l \times W_l$. To produce the final uncertainty map, we first upsample each $\mathbf{M}_l$ to the output dimensions via bicubic interpolation, then aggregate using resolution-weighted averaging:
\begin{equation}
    \mathbf{M}_{\text{RADMI}} = \sum_{l=1}^{L-1} w_l \cdot \text{Upsample}(\mathbf{M}_l), \quad w_l = \frac{H_l \times W_l}{\sum_{l'=1}^{L-1} H_{l'} \times W_{l'}}
\end{equation}

\begin{figure}[h]
\begin{minipage}[b]{\linewidth}
  \centering
  \centerline{\includegraphics[width=8.5cm]{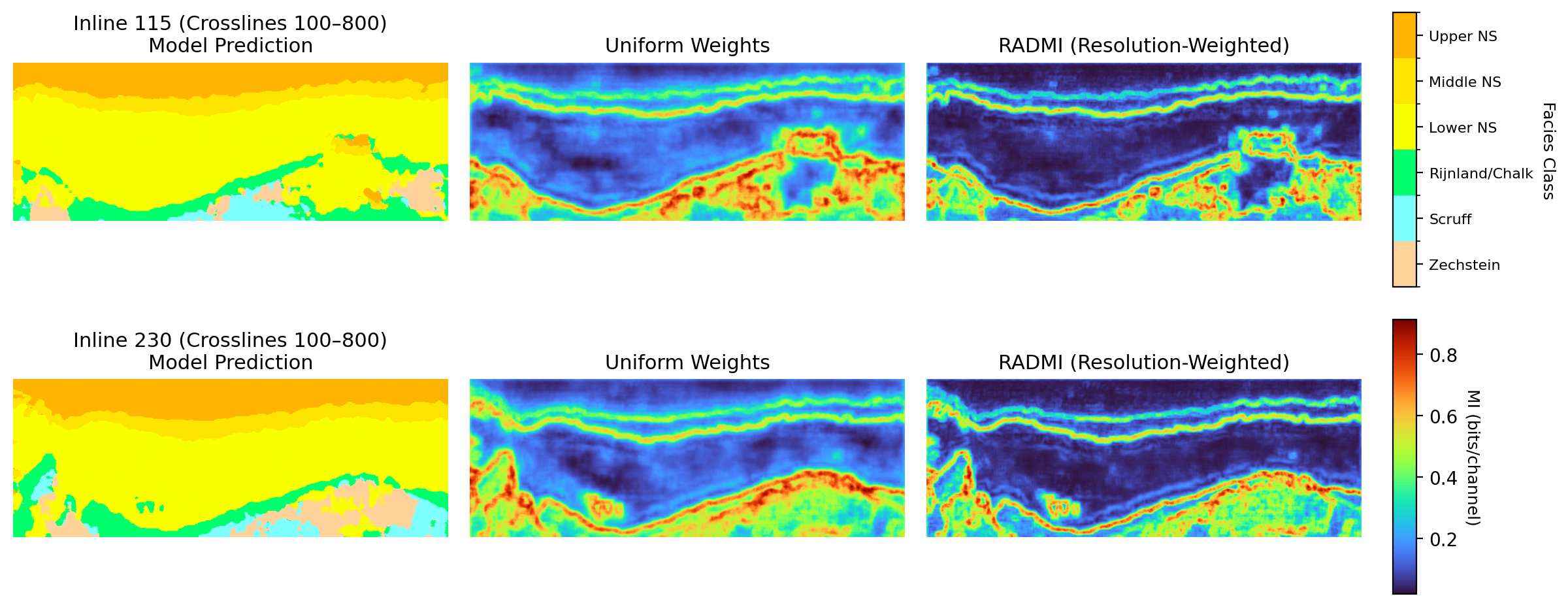}}
%  \vspace{2.0cm}
\end{minipage}
\vspace{-10mm}
\caption{Decoder MI Weighting Ablation.}
\label{fig:ablation}
\end{figure}

This weighting scheme emphasizes MI estimates computed at finer resolutions, where spatial localization is more precise. Figure~\ref{fig:ablation} compares resolution-weighted aggregation against uniform weighting ($w_l = 1/(L-1)$). Resolution weighting produces sharper, more localized boundary responses compared to uniform weighting, which yields broader and more diffuse uncertainty regions.

\vspace{-2mm}

%% file: sections/05-experiments.tex
\begin{figure*}[htb]
\vspace{-10mm}
\begin{minipage}[b]{1.0\linewidth}
  \centering
  \centerline{\includegraphics[width=17cm]{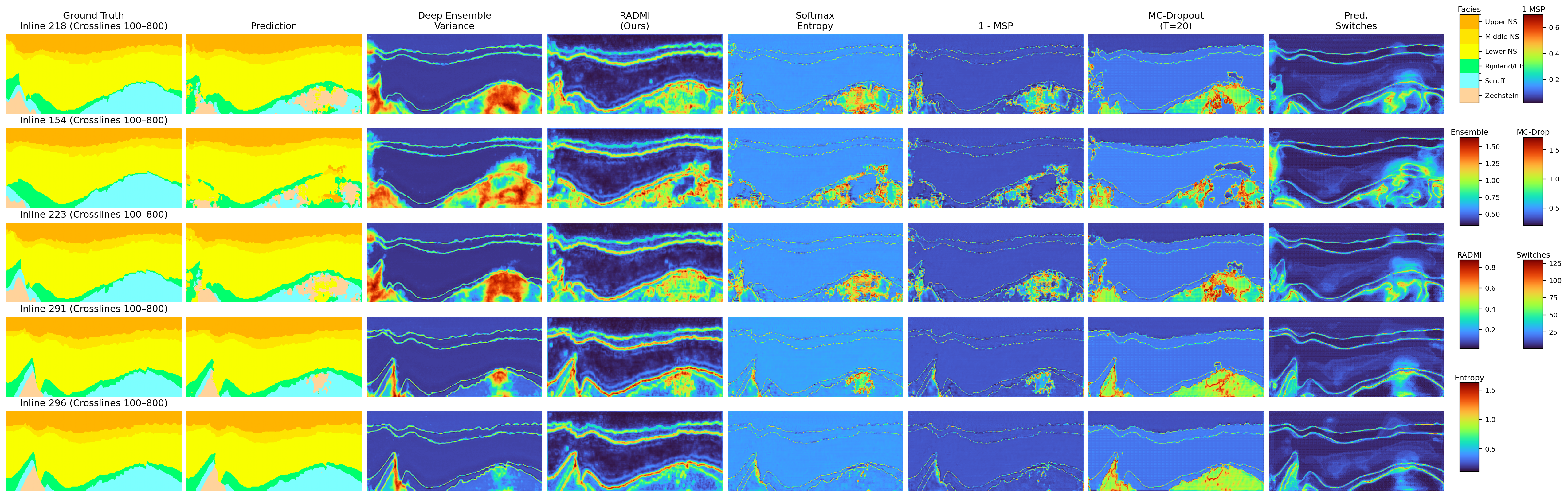}}
%  \vspace{2.0cm}
\end{minipage}
\vspace{-5mm}
\caption{Examples of Uncertainty Heatmaps from Various Method Applied to Facies Segmentation.}
\label{fig:result}
\end{figure*}

\subsection{Experimental Setup}

We evaluate our proposed \texttt{RADMI} method on the F3 seismic facies segmentation benchmark \cite{alaudahfacies}, 
a widely-used dataset for seismic interpretation containing 6 lithological classes: Upper North Sea, Middle North Sea, Lower North Sea, Rijnland/Chalk, Scruff, and Zechstein. The F3 block, located in the Dutch sector of the North Sea, presents challenges including gradual facies transitions and complex stratigraphic boundaries. 

Our base architecture is FaciesSegNet, a U-Net variant with encoder channel dimensions $(8, 10, 30, 40, 60)$ employing batch normalization after each convolutional block. The network is trained on 360 inline sections with a 90/10 train/validation split, using the Adam optimizer with learning rate $10^{-3}$, and weight decay $10^{-5}$. All experiments are conducted on an independent test set of 200 seismic sections that exhibit natural domain shift from the training volume.

\subsubsection{Baseline Methods}

We compare \texttt{RADMI} against established uncertainty quantification methods spanning both single-model and multi-model paradigms. First, we implement a deep ensemble. We train $M=30$ independently initialized models with different random seeds. Ensemble uncertainty is computed as the predictive entropy of the averaged softmax probabilities across members. Next, we obtain logit-based uncertainty. This includes softmax entropy:  the standard predictive entropy $H = -\sum_c p_c \log p_c$ computed directly from the softmax output distribution; and 1 - Max Softmax Probability (MSP): a confidence-based uncertainty measure $U = 1 - \max_c p_c$ that quantifies how far the network's most confident prediction falls from certainty. We further implement the methods of prediction switch tracking \cite{benkert2022reliableuncertaintyestimation} and MC-Dropout \cite{galpmlr}. Prediction switch tracking counts how frequently each pixel's predicted class changes across training epochs, providing a discrete measure of disagreement. Pixels with high switch counts indicate regions where the frequently changes the prediction. The MC-Dropout implementation follows the setup in \cite{benkert2022reliableuncertaintyestimation}.
We train a separate model with dropout layers ($p=0.5, 0.1$) inserted in the final two decoder blocks. At inference, we perform $T=20$ stochastic forward passes with dropout enabled and compute predictive entropy from the averaged softmax predictions. This approximates Bayesian inference over network weights.

\subsubsection{Evaluation Process}

Since ground-truth uncertainty is unavailable, we follow the common practice of treating deep ensemble uncertainty as a reference standard. We evaluate all methods by measuring their agreement with ensemble predictive entropy across multiple metrics. This evaluation framework is motivated by the observation that ensembles capture both epistemic and aleatoric uncertainty through model diversity, and if a single-pass method can approximate ensemble uncertainty without the computational overhead, it provides practical value for real-world scenarios.

We report three categories of metrics: correlation metrics, distance metrics, and overlap metrics. Pearson correlation, Spearman rank correlation, and cosine similarity measure linear, monotonic, and directional agreement respectively. KL divergence, Jensen-Shannon divergence, $L_2$ distance, bidirectional Chamfer distance, and Earth Mover's distance (EMD) quantify distributional and spatial discrepancies. Mean IoU (computed across multiple thresholds), DICE coefficient, and histogram intersection assess the spatial and distributional overlap between uncertainty maps.

All metrics are computed per-section and aggregated across the 200 test sections, reporting mean $\pm$ standard deviation.

\subsection{Qualitative Results}

Figure~\ref{fig:result} presents uncertainty maps from all evaluated methods on five representative test sections. \texttt{RADMI} produces uncertainty maps with sharp, well-localized responses at facies boundaries. The deep ensemble similarly highlights these boundary regions. In contrast, softmax entropy and 1-MSP yield more diffuse uncertainty patterns that spread across large portions of each section, failing to distinguish boundaries from interior regions. The 1-MSP maps in particular show elevated uncertainty throughout the shallower formations regardless of proximity to class boundaries.

\texttt{RADMI} exhibits strong visual correspondence with the deep ensemble at regions of prediction difficulty. MC-Dropout captures some of these structures but with less spatial precision, while softmax entropy shows only weak response. Prediction switches produces the finest boundary delineation among all methods, with very sharp transitions between certain and uncertain regions. However, this sharpness comes at the cost of a fundamentally different uncertainty distribution. The discrete counting nature of switches produces a coarser value range compared to the continuous entropy-based measures. Despite this distributional difference, the spatial agreement between prediction switches and \texttt{RADMI} at boundary locations supports the interpretation that both capture semantically meaningful uncertainty signals through different mechanisms.

\begin{table*}[htb]
    \vspace{-10mm}
    \centering
    
    \resizebox{0.9\textwidth}{!}{%
    \begin{tabular}{l|ccc|ccc}
        \hline
        & \multicolumn{3}{c}{\textbf{Correlation Metrics} $\uparrow$} & \multicolumn{3}{c|}{\textbf{Overlap Metrics} $\uparrow$} \\
        \cline{2-7}
        \textbf{Method} & Pearson & Spearman & Cosine & mIoU & DICE & Hist. Int. \\
        \hline
        Softmax Entropy~\cite{hendrycks2016baseline} & $0.605 \pm 0.043$ & $0.230 \pm 0.061$ & $\mathbf{0.926 \pm 0.015}$ & $0.298 \pm 0.023$ & $0.412 \pm 0.028$ & $0.227 \pm 0.038$ \\
        1 - MSP~\cite{hendrycks2016baseline} & $0.571 \pm 0.038$ & $0.239 \pm 0.060$ & $0.853 \pm 0.022$ & $0.239 \pm 0.018$ & $0.337 \pm 0.023$ & $0.304 \pm 0.062$ \\
        MC-Dropout~\cite{galpmlr} ($T$=20) & $0.639 \pm 0.051$ & $0.273 \pm 0.054$ & $0.923 \pm 0.012$ & $0.320 \pm 0.033$ & $\mathbf{0.437 \pm 0.039}$ & $0.370 \pm 0.061$ \\
        Pred. Switches~\cite{benkert2022reliableuncertaintyestimation} & $0.627 \pm 0.063$ & $0.505 \pm 0.085$ & $0.818 \pm 0.016$ & $0.228 \pm 0.019$ & $0.308 \pm 0.025$ & $0.461 \pm 0.048$ \\
        \textbf{RADMI (Ours)} & $\mathbf{0.674 \pm 0.041}$ & $\mathbf{0.559 \pm 0.086}$ & $0.861 \pm 0.019$ & $\mathbf{0.325 \pm 0.037}$ & $0.427 \pm 0.038$ & $\mathbf{0.472 \pm 0.065}$ \\
        \hline
    \end{tabular}%
    }
    
    \vspace{1mm}
    
    \resizebox{0.75\textwidth}{!}{%
    \begin{tabular}{l|ccccc}
        \hline
        & \multicolumn{5}{c}{\textbf{Distance Metrics} $\downarrow$} \\
        \cline{2-6}
        \textbf{Method} & KL Div & JS Div & $L_2$ Dist & Chamfer & EMD \\
        \hline
        Softmax Entropy~\cite{hendrycks2016baseline} & $0.358 \pm 0.064$ & $0.292 \pm 0.022$ & $0.887 \pm 0.048$ & $\mathbf{2.903 \pm 0.779}$ & $\mathbf{0.003 \pm 0.001}$ \\
        1 - MSP~\cite{hendrycks2016baseline} & $0.328 \pm 0.059$ & $0.286 \pm 0.024$ & $0.926 \pm 0.041$ & $3.450 \pm 0.759$ & $0.005 \pm 0.001$ \\
        MC-Dropout~\cite{galpmlr} ($T$=20) & $\mathbf{0.309 \pm 0.072}$ & $0.270 \pm 0.027$ & $0.847 \pm 0.059$ & $3.290 \pm 1.351$ & $0.004 \pm 0.001$ \\
        Pred. Switches~\cite{benkert2022reliableuncertaintyestimation} & $0.312 \pm 0.055$ & $0.267 \pm 0.023$ & $0.860 \pm 0.074$ & $6.576 \pm 2.766$ & $0.006 \pm 0.001$ \\
        \textbf{RADMI (Ours)} & $0.339 \pm 0.031$ & $\mathbf{0.265 \pm 0.013}$ & $\mathbf{0.806 \pm 0.048}$ & $5.132 \pm 1.899$ & $0.008 \pm 0.001$ \\
        \hline
    \end{tabular}%
    }
    \vspace{-2mm}
    \caption{Agreement with Deep Ensemble Uncertainty.}
    \label{tab:ensemble_agreement}
\end{table*}

\subsection{Quantitative Evaluation}

\subsubsection{Agreement with Ensemble Uncertainty}

Table~\ref{tab:ensemble_agreement} presents correlation metrics measuring agreement with deep ensemble uncertainty. \texttt{RADMI} achieves the highest Pearson correlation among all single-pass methods, outperforming softmax entropy, 1-MSP, and MC-Dropout. This indicates that MI-based uncertainty captures the linear relationship with ensemble uncertainty more effectively than output-layer methods.

The Spearman correlation results are particularly noteworthy. \texttt{RADMI} performs substantially higher than softmax entropy, 1-MSP, and MC-Dropout. Spearman correlation measures monotonic agreement rather than linear agreement, making it robust to differences in scale and distribution shape. The strong Spearman performance suggests that \texttt{RADMI} uncertainty correctly ranks pixels by uncertainty level, even if the absolute values differ from ensemble entropy. This is important for real-world applications where relative uncertainty ordering matters. Prediction switches also achieves high Spearman correlation, which aligns with its design as a rank-based disagreement measure. However, prediction switches requires multiple trained models, whereas \texttt{RADMI} operates on a single forward pass. The cosine similarity results reveal an interesting trade-off. Softmax entropy achieves the highest cosine similarity, followed by MC-Dropout, while \texttt{RADMI} is slightly lower. Cosine similarity measures directional alignment in the high-dimensional pixel space and is insensitive to magnitude differences. The high cosine similarity of softmax methods suggests they capture similar patterns of uncertainty to the deep ensemble, but may be different in magnitude or ranking. \texttt{RADMI}'s lower cosine similarity but higher Pearson and Spearman correlations indicates it better captures both the magnitude and rank structure of ensemble uncertainty, at the cost of some pattern-level agreement.

Table~\ref{tab:ensemble_agreement} also presents overlap and distance metrics comparing each method's uncertainty maps to deep ensemble uncertainty. \texttt{RADMI} achieves the lowest $L_2$ distance and Jensen-Shannon divergence among all methods, indicating strong distributional agreement with ensemble uncertainty in terms of overall magnitude and spread. MC-Dropout achieves the lowest KL divergence, with \texttt{RADMI} close behind. The Chamfer distance and EMD results reflect a fundamental difference in how \texttt{RADMI} represents uncertainty compared to softmax-based methods. \texttt{RADMI} produces sharp, boundary-concentrated uncertainty maps, whereas ensemble uncertainty exhibits broader spatial extent around class transitions. When these maps are binarized for Chamfer computation, \texttt{RADMI}'s thin boundary responses yield higher average distances to the ensemble's broader regions even though the same semantic locations are captured. Similarly, the higher EMD reflects \texttt{RADMI}'s more bimodal value distribution (low in interiors, high at boundaries) compared to the more continuous distribution of ensemble entropy. These metrics penalize distributional shape differences rather than spatial or rank agreement, which the correlation metrics better capture.

\texttt{RADMI} achieves the highest mIoU across all methods, indicating that when uncertainty maps are thresholded at multiple levels, \texttt{RADMI}'s high-uncertainty regions best overlap with those of the ensemble. \texttt{RADMI} also achieves the highest histogram intersection, demonstrating substantial agreement in value distributions despite the EMD results. MC-Dropout achieves marginally higher DICE. Notably, softmax entropy and 1-MSP show considerably lower overlap scores across all three metrics, consistent with the qualitative observation that these methods fail to localize uncertainty in full, only capturing central areas of ensemble uncertainty rather than the full structure.

\subsubsection{Agreement with Prediction Error}

To further validate the use of deep ensemble uncertainty as the reference standard, we compare all methods directly against prediction error maps (pixels where the model prediction disagrees with the ground truth). Table~\ref{tab:error} presents correlation and overlap metrics against prediction error.

\begin{table}[htb]
    \centering
    
    \resizebox{\columnwidth}{!}{%
    \begin{tabular}{l|ccc}
        \hline
        & \multicolumn{3}{c}{\textbf{Correlation Metrics} $\uparrow$} \\
        \cline{2-4}
        \textbf{Method} & Pearson & Spearman & Cosine \\
        \hline
        Deep Ensemble~\cite{lakshminarayanan2017simple} & $\mathbf{0.715 \pm 0.081}$ & $\mathbf{0.568 \pm 0.108}$ & $\mathbf{0.661 \pm 0.114}$ \\
        \textbf{RADMI (Ours)} & $0.543 \pm 0.074$ & $0.496 \pm 0.085$ & $0.634 \pm 0.095$ \\
        Softmax Entropy~\cite{hendrycks2016baseline} & $0.493 \pm 0.056$ & $0.291 \pm 0.063$ & $0.549 \pm 0.093$ \\
        1 - MSP~\cite{hendrycks2016baseline} & $0.467 \pm 0.051$ & $0.297 \pm 0.063$ & $0.582 \pm 0.077$ \\
        MC-Dropout~\cite{galpmlr} ($T$=20) & $0.467 \pm 0.073$ & $0.382 \pm 0.069$ & $0.559 \pm 0.096$ \\
        Pred. Switches~\cite{benkert2022reliableuncertaintyestimation} & $0.462 \pm 0.061$ & $0.437 \pm 0.060$ & $0.588 \pm 0.064$ \\
        \hline
    \end{tabular}%
    }
    
    \vspace{1mm}
    
    \resizebox{0.8\columnwidth}{!}{%
    \begin{tabular}{l|cc}
        \hline
        & \multicolumn{2}{c}{\textbf{Overlap Metrics} $\uparrow$} \\
        \cline{2-3}
        \textbf{Method} & mIoU & DICE \\
        \hline
        Deep Ensemble~\cite{lakshminarayanan2017simple} & $\mathbf{0.378 \pm 0.079}$ & $\mathbf{0.506 \pm 0.086}$ \\
        \textbf{RADMI (Ours)} & $0.234 \pm 0.051$ & $0.341 \pm 0.062$ \\
        Softmax Entropy~\cite{hendrycks2016baseline} & $0.207 \pm 0.033$ & $0.312 \pm 0.043$ \\
        MC-Dropout~\cite{galpmlr} ($T$=20) & $0.207 \pm 0.037$ & $0.318 \pm 0.051$ \\
        1 - MSP~\cite{hendrycks2016baseline} & $0.179 \pm 0.026$ & $0.274 \pm 0.033$ \\
        Pred. Switches~\cite{benkert2022reliableuncertaintyestimation} & $0.151 \pm 0.021$ & $0.231 \pm 0.029$ \\
        \hline
    \end{tabular}%
    }
    \vspace{-2mm}
    \caption{Agreement with Prediction Error.}
    \label{tab:error}
    %\vspace{-5mm}
\end{table}

Deep ensemble achieves the highest correlation with prediction error across all metrics, providing empirical justification for treating it as the reference standard in our primary evaluation. Among single-pass methods, \texttt{RADMI} achieves the highest Pearson correlation, outperforming MC-Dropout by 16\% and softmax entropy by 10\%. \texttt{RADMI} also achieves the highest Spearman correlation among single-pass methods, a 30\% improvement over MC-Dropout. The overlap metrics show a similar pattern: deep ensemble achieves the best mIoU and DICE, with \texttt{RADMI} ranking second among all methods. These results demonstrate that \texttt{RADMI} not only approximates ensemble uncertainty well, but also captures meaningful signal about where the model actually makes errors.

\subsubsection{Computational Cost}

Another key advantage of \texttt{RADMI} is computational efficiency. Table~\ref{tab:compute} summarizes the requirements of each method. Single-pass methods (softmax entropy, 1-MSP, and \texttt{RADMI}) require only one forward pass through the network. \texttt{RADMI} incurs a low overhead for computing pairwise MI between decoder layers, requiring just a single forward pass. This is much less than the other competitive methods, such as the $20\times$ overhead of MC-Dropout, the $30\times$ overhead of deep ensembles, and the $300\times$ overhead prediction switches.

\begin{table}[htb]
    \centering
    \caption{Computational requirements for uncertainty estimation on new data.}
    \vspace{1mm}
    \label{tab:compute}
    \resizebox{\columnwidth}{!}{%
    \begin{tabular}{lccc}
        \hline
        \textbf{Method} & \textbf{Training Cost} & \textbf{Forward Passes}\\
        \hline
        Softmax Entropy~\cite{hendrycks2016baseline} & $1\times$ & 1 \\
        1 - MSP~\cite{hendrycks2016baseline} & $1\times$ & 1 \\
        MC-Dropout~\cite{galpmlr} ($T$=20) & $1\times$ & 20 \\
        Deep Ensemble~\cite{lakshminarayanan2017simple} ($M$=30) & $30\times$ & 30 \\
        Pred. Switches~\cite{benkert2022reliableuncertaintyestimation} & $1\times$ & 300* & \\
        \textbf{RADMI (Ours)} & $1\times$ & 1 \\
        \hline
    \end{tabular}
    }
    \vspace{1.5mm}
    \footnotesize{*Prediction Switches require saving model checkpoints at every epoch.}
    %\vspace{-5mm}
\end{table}

For real-time or large-scale seismic interpretation workflows spanning hundreds of square kilometers worth of data, this efficiency advantage is significant. \texttt{RADMI} requires no ensemble training (which multiplies training cost by $M$), no architectural modifications for dropout, and no repeated inference, making it practical for use in real-world situations.

%% file: sections/06-conclusion.tex
We have presented \texttt{RADMI}, a single-pass uncertainty estimation method based on mutual information between consecutive decoder layers. By directly measuring inter-layer information flow, \texttt{RADMI} outperforms the next-best baselines by 5.5\% in Pearson and 10.7\% in Spearman correlation with deep ensemble uncertainty, while producing sharp, boundary-localized maps suitable for identifying regions requiring expert attention. The approach requires no architectural modifications and is agnostic to the specific encoder-decoder architecture, suggesting applicability beyond seismic interpretation to other dense prediction domains. Future work includes extension to 3D volumetric data and validation on medical imaging and remote sensing tasks.